\documentclass[11pt]{article}

\usepackage[T1]{fontenc}
\usepackage[utf8]{inputenc}
\usepackage{lmodern}
\usepackage{microtype}
\usepackage[a4paper,margin=1in]{geometry}
\usepackage{graphicx}
\usepackage{booktabs}
\usepackage{subfigure}
\usepackage{xcolor}
\usepackage{hyperref}
\usepackage{amsmath,amssymb}
\usepackage{adjustbox}
\usepackage{multirow}
\usepackage{tcolorbox}
\tcbuselibrary{breakable}
\usepackage{algorithm}
\usepackage{algpseudocode}
\usepackage{changepage}
\usepackage[numbers,sort&compress]{natbib}

\definecolor{lightblu}{RGB}{229,240,250}
\definecolor{lightgrey}{RGB}{240,240,240}
\definecolor{lightyellow}{RGB}{255,255,224}
\definecolor{lightgreen}{RGB}{215,238,145}
\definecolor{lightpurple}{RGB}{230,230,250}
\definecolor{lightorange}{RGB}{255,200,61}

\hypersetup{
  colorlinks=true,
  linkcolor=blue,
  citecolor=blue,
  urlcolor=blue,
  pdfauthor={Fabrizio Marozzo and Stefano Iannicelli},
  pdftitle={Large Language Models for Token-Efficient and Semantic-Preserving Opinion Summarization}
}

\title{Large Language Models for Token-Efficient and Semantic-Preserving Opinion Summarization}
\author{
Fabrizio Marozzo\\
University of Calabria, Rende, Italy\\
\texttt{fmarozzo@dimes.unical.it}
\and
Stefano Iannicelli\\
University of Calabria, Rende, Italy
}
\date{}

\begin{document}
\maketitle

\begin{abstract}
Opinionated text---spanning product reviews, hotel feedback, and social posts---captures rich signals about user experiences, preferences, and concerns. However, the scale, redundancy, and imbalance of such corpora make it challenging to analyze opinions effectively, particularly when the goal is to generate summaries that remain faithful to the diversity of viewpoints expressed. This paper presents a framework that preserves semantics in LLM-based opinion summarization while minimizing token usage. We combine multidimensional classification (e.g., sentiment, topics) with a family of stratified sampling strategies to select compact yet representative subsets of opinions before prompting the LLM. Tailored prompts then produce balanced summaries that surface the salient aspects expressed in the opinions (e.g., strengths and weaknesses of products/hotels). Experiments on Amazon product reviews, Tripadvisor hotel reviews, and X/Twitter posts demonstrate that our method significantly reduces token usage and computational cost while consistently outperforming traditional AI-based and standard LLM summarization baselines in terms of content coverage, balance, and semantic preservation.
\end{abstract}


\section{Introduction}
\label{sec:intro}

User-generated opinions in the form of product reviews, hotel assessments, social posts, and peer-support discussions constitute a rich source of information for capturing user experiences, preferences, and concerns \cite{yang2019exploiting,zhang2014examining}. Organizations rely on these signals to evaluate product performance, identify service issues, and monitor emerging trends, while users consult them to gather information, compare alternatives, and make informed decisions \cite{liu2005opinion}. Yet the scale, heterogeneity, and redundancy of these corpora make effective opinion analysis challenging, especially when one aims to preserve the diversity and nuance of the viewpoints expressed \cite{pang2002thumbs,sun2023text}.

Large Language Models (LLMs) offer strong capabilities for opinion classification and analysis, but applying them directly to full corpora is often inefficient and prone to bias: context windows are finite, the cost of processing large inputs is high, and model outputs tend to overweight majority viewpoints when the input distribution is not controlled \cite{roumeliotis2024llms}. Existing pipelines frequently rely on single-dimension analysis (e.g., sentiment only \cite{sun2023text,roumeliotis2024llms}), fail to account for the need to balance multiple semantic facets such as topics and emotions, or lose important information when opinion sets are reduced without principled criteria. These limitations highlight the need for methods that preserve the diversity and structure of opinionated content while supporting efficient large-scale analysis.

This paper introduces a framework designed to preserve semantics in LLM-based opinion summarization while minimizing input size. The approach first applies multidimensional classification---such as sentiment, topics, emotion, and optional domain-specific facets---to impose structure on the corpus. The framework then employs stratified sampling strategies to select compact yet representative subsets of opinions before prompting the LLM. By feeding the model a subset that is already balanced across facets and rich in informative content, the LLM can generate summaries that are both more faithful and less biased than those obtained from raw or arbitrarily reduced inputs. Tailored prompts guide the model to recover the salient aspects expressed in the corpus, such as product strengths and weaknesses, hotel service issues, and political support rationales. Unlike RAG-based approaches \cite{lewis2020rag,gao2023rag_survey} or methods that rely on ever-larger context windows, our framework optimizes what is fed to the model: a facet-balanced, corpus-level selection that improves generation quality while avoiding the cost and skew trade-offs of brute-force ingestion or query-specific retrieval. To support reproducibility, the implementation of the proposed framework is publicly available at \url{https://github.com/SCAlabUnical/semantic-preserving-summarization}.

We evaluate our framework on Amazon product reviews, Tripadvisor hotel evaluations, and political discussions on Twitter/X, configuring facet sets to match each domain. Across all corpora, our framework substantially reduces the number of tokens required for summarization while maintaining high content coverage, facet balance, and semantic preservation compared with standard LLM summarization. By providing the LLM with a compact yet distribution-aligned subset, the model produces summaries that more accurately reflect the diversity of opinions rather than overemphasizing majority viewpoints. These results demonstrate that token-efficient, distribution-aligned selection combined with facet-aware prompting yields compact yet faithful summaries, supporting scalable opinion analysis across heterogeneous domains.

This article substantially extends our previous work~\cite{ecml-pkdd-marozzo-2025}, which introduced a framework for generating balanced summaries of user reviews by selecting representative opinions from multidimensional classification results before passing them to a Large Language Model (LLM) to reduce input tokens while preserving opinion diversity. Compared with the conference version, this journal article introduces five main advances:

\begin{itemize}
\item \textit{Broader Algorithmic Scope}: Unlike the conference version, which considered a narrower balanced-selection setting, we introduce and compare three distinct distribution-aware sampling techniques for constructing balanced subsets before summarization: a relevance-constrained \texttt{Knapsack} strategy, a KL-regularized \texttt{Knapsack-KL} variant, and a density-based \texttt{KDE} strategy.

\item \textit{More Rigorous Algorithmic Formulation}: We provide a formal treatment of the proposed selection strategies, clarifying their design principles, computational properties, and their role in preserving representative opinions across sentiment, topic, and other semantic facets.

\item \textit{Expanded Empirical Validation}: We extend the experimental analysis to multiple datasets and domains, allowing a more robust assessment of how different sampling strategies affect semantic preservation, topic coverage, and summary fidelity.

\item \textit{Task-Aligned Evaluation Protocol}: Unlike the conference version, which primarily emphasized overall summary quality through lexical, semantic, and qualitative measures, the present article adopts an evaluation protocol specifically designed to assess sampling strategies under token constraints, focusing on topic coverage, summary-level cosine similarity, and token usage.

\item \textit{Deeper Efficiency Analysis}: We examine the trade-offs among subset size, token usage, computational cost, and semantic preservation, and provide additional analyses to clarify how different selection strategies affect the balance, representativeness, and interpretability of the generated summaries.
\end{itemize}

The remainder of this paper is structured as follows. Section~\ref{sec:related} reviews related work. Section~\ref{sec:framework} presents the proposed framework. Section~\ref{sec:experiments} discusses the results. Finally, Section~\ref{sec:conclusions} concludes the paper.

\section{Related work}\label{sec:related}

Large Language Models (LLMs) have been widely adopted across domains such as education, e-commerce, healthcare, and cybersecurity for tasks including question answering, report generation, and data visualization \cite{info13010041,okonkwo2021chatbots}. In parallel, automatic summarization has become increasingly important for condensing large collections of text into compact and decision-oriented representations, especially when the source material is redundant, noisy, and difficult to inspect manually \cite{Chengyao2025survey,sharma2022automatic}. This need is particularly evident in opinion-rich corpora, where effective summaries should preserve not only the dominant topics, but also the diversity of viewpoints and attitudes expressed across documents \cite{10.1145/3477495.3532676}. In practical terms, high-quality and balanced summarization is valuable because it helps consumers make informed purchasing decisions \cite{ecml-pkdd-marozzo-2025}, supports comparisons among competing products and services based on user reviews \cite{product-comp-ds25}, and enables organizations to identify recurring strengths, weaknesses, and unmet user needs from large collections of authentic feedback \cite{novelneeds2024,reviewsproductdev2020}.

In opinion mining, LLMs and neural methods have substantially improved topic modeling, sentiment and emotion analysis, and summarization \cite{cantini2025harnessing,ROUMELIOTIS2024100056}, building on classical sentiment-analysis research \cite{pang2002thumbs,liu2005opinion} and more recent advances for tweets and product feedback \cite{anbumani2024enhancing,10.1145/3615356}. However, summarizing full corpora remains token-inefficient and may overweight majority viewpoints, which motivates pipelines that structure and filter inputs before generation \cite{sun2023text,roumeliotis2024llms}. Existing approaches often rely on single-dimension filtering---typically sentiment-only---and therefore do not adequately balance multiple semantic facets such as topics and emotions, limiting their ability to produce representative summaries of heterogeneous opinion spaces.

Beyond single-dimension filtering, a related line of research focuses on selecting representative and diverse subsets of documents before summarization. In multi-document and opinion summarization, this problem is particularly important because the quality of the final summary depends not only on the generation model, but also on whether the selected inputs adequately cover the main aspects and viewpoints present in the corpus \cite{carbonell1998use,zhang2015clustering}. Existing approaches have explored clustering-based selection, diversity-promoting ranking, and coverage-oriented sampling \cite{jiang2023large,carichon2023topically}, but they often emphasize lexical or semantic redundancy reduction without explicitly preserving multidimensional opinion distributions. Our work is aligned with this direction, but differs in that it uses structured semantic facets---such as sentiment, topic, and emotion---to guide subset construction under explicit token constraints.

Retrieval-Augmented Generation (RAG) mitigates some limitations by grounding outputs in retrieved evidence, improving factuality and reducing hallucination \cite{lewis2020rag}. RAG variants differ in retriever design, indexing, reranking, and fusion mechanisms \cite{gao2023rag_survey}. More recent practice trends toward ever-larger context windows to ingest extensive portions of a corpus, but this increases token, compute, and memory costs and may amplify majority-view bias. Our work moves in the opposite direction: rather than ingesting or retrieving large volumes of text, we perform corpus-level, facet-aware selection to construct a small, stratified, and representative subset before generation. This complements RAG by optimizing what is fed to the model, not only how evidence is retrieved.

Automatic evaluation has moved beyond simple n-gram overlap metrics such as ROUGE \cite{lin2004rouge} toward embedding-based measures like BERTScore \cite{zhang2019bertscore}, which better capture semantic similarity. Fairness-oriented work in multi-document summarization has also emerged, introducing metrics to assess balanced representation across groups \cite{li2025coverage-fairness}. While these metrics provide useful perspectives, they are not directly aligned with our goal of preserving semantic structure under strict token budgets. Instead, we evaluate our method through topic-coverage analysis and summary-level cosine similarity, which more precisely quantify how well selected subsets reflect the full corpus and how faithfully the resulting summaries preserve meaning.

\section{Framework}
\label{sec:framework}

\begin{figure}[tb!] 
\begin{adjustwidth}{-1cm}{1cm} 
\centering \includegraphics[width=1.2\linewidth]{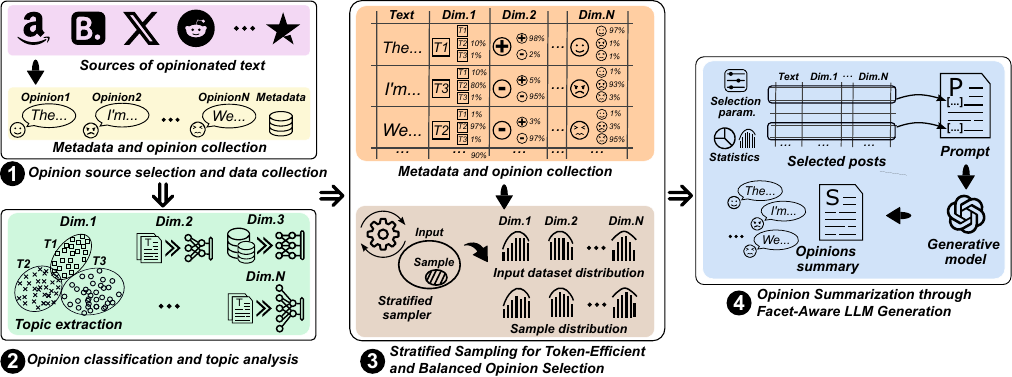} \caption{Execution flow of the proposed framework.} \label{fig:proposed_framework} 
\end{adjustwidth} 
\end{figure}

We now describe the proposed framework, which transforms raw opinionated text into compact, semantically balanced inputs for LLM-based summarization. As shown in Figure~\ref{fig:proposed_framework}, the system operates in four stages: (i) data collection, (ii) multidimensional classification and topic analysis, (iii) stratified sampling, and (iv) facet-aware summarization.

In the \textit{first stage}, the system collects opinions from heterogeneous domains such as \emph{Amazon} product reviews, \emph{Tripadvisor} hotel evaluations, and \emph{Twitter/X} political posts. Data are obtained through official APIs or custom crawlers, together with metadata (e.g., timestamps, ratings, helpfulness votes). Optional filters—such as keywords, time windows, or category constraints—ensure that the collected opinions are relevant and representative.

The \textit{second stage} performs \emph{opinion classification and topic extraction}. Each opinion is annotated across multiple semantic dimensions, including \emph{sentiment}, \emph{emotion}, and \emph{topics}, plus optional \emph{domain-specific facets} such as candidate support. For each dimension, a corresponding set of classes is defined (e.g., positive/negative sentiment, anger/joy emotion, dataset-specific topics). Transformer-based classifiers assign probabilistic scores rather than discrete labels, while BERTopic provides embedding-based topic discovery. This results in a structured, probabilistic representation that supports balanced downstream selection (full details in Section~\ref{sec:multi-classification}).

The \textit{third stage} applies \emph{stratified sampling} to select a compact yet representative subset of opinions under a given token budget. The goal is to preserve class distributions across all semantic dimensions while retaining diverse and information-rich instances. Multiple strategies (Knapsack, Kullback–Leibler, KDE) are discussed in Section~\ref{sec:sampling}.

Finally, the \textit{last stage} generates concise and semantically rich summaries from the sampled subset. Prompts explicitly request coverage of the relevant facets identified during classification, for example, strengths and weaknesses of products or hotels, or balanced political arguments. By combining structured classification with distribution-preserving sampling, the framework yields summaries that are both compact and faithful, achieving substantial token savings without sacrificing semantic quality.

\subsection{Multi-Dimensional Classification of Opinions}
\label{sec:multi-classification}

We enrich each opinion with semantic attributes derived from transformer-based classifiers and BERTopic. Each text instance is mapped to a multidimensional representation composed of probabilistic scores across several dimensions. Across all case studies, we model: \texttt{Sentiment} (classes: \textit{positive}, \textit{negative}), \texttt{Emotion} (classes: \textit{anger}, \textit{disgust}, \textit{fear}, \textit{joy}, \textit{neutral}, \textit{sadness}, \textit{surprise}), \texttt{Topic} (classes: \textit{dataset-specific topics}), with optional domain-specific facets such as \texttt{candidate support} (e.g., \textit{pro-A} / \textit{pro-B}).

For each dimension, we select a dedicated classifier following a model comparison procedure consistent with prior studies~\cite{zhang2024survey,cantini2024multi}. We focus on transformer-based architectures for their stability and efficient inference; prompt-based generative classifiers were evaluated but discarded due to stochastic behaviour. Sentiment is modeled using BERT, fine-tuned on annotated hotel reviews.\footnote{\url{https://www.kaggle.com/datasets/jiashenliu/515k-hotel-reviews-data-in-europe}} Emotion is identified using a BERT multi-label classifier fine-tuned on SemEval-2018.\footnote{\url{https://huggingface.co/ayoubkirouane/BERT-Emotions-Classifier}} Topics are extracted using BERTopic~\cite{grootendorst2022bertopic}, which provides coherent and diverse themes across domains~\cite{egger2022topic}. 
For domain-specific facets, we adopt the best-performing model for each specific dataset (e.g., RoBERTa for candidate-support classification), using the same evaluation criteria as for the core dimensions.

The resulting multidimensional, probabilistic representation enables fine-grained control over semantic coverage, supporting the stratified sampling strategies and improving the model’s ability to generate balanced, facet-aware summaries.

\subsection{Stratified Sampling Strategies}
\label{sec:sampling}

Given a corpus of opinions, each instance is represented by probabilistic labels across one or more semantic dimensions (e.g., sentiment, topics, emotion, and optionally domain-specific facets such as candidate support or psychological distress). For each dimension \(d\) with class set \(C(d)\), an opinion \(r\) is associated with a probability vector \(P(c \mid r, d)\) over \(c \in C(d)\). Our goal is to select a subset \(S\) of fixed size \(N\) that: (i) preserves the class distributions of the full corpus across all selected dimensions, and (ii) maximizes the information density of the input passed to the LLM under a strict token budget. We consider three different stratified sampling strategies. 

The \texttt{Knapsack} strategy casts selection as a constrained optimization problem. First, for each dimension \(d\) and class \(c \in C(d)\), we estimate the target proportion \(P(c \mid d)\) in the full corpus and derive a target count \(N_{c \mid d}\) for the subset. Opinions are then processed in descending order of relevance score, and each one is added to \(S\) only if it does not violate any of the per-class targets. Intuitively, this behaves like a multi-dimensional knapsack: relevance plays the role of item value, while the target counts act as capacity constraints. The result is a subset that is both informative and approximately distribution-preserving across all considered dimensions.

The relevance score measures how strongly a review reflects the main topics identified by BERTopic. It is calculated by counting the occurrences of the top-k BERTopic terms in the review and normalizing it by the highest such count across all reviews, producing a score between 0 and 1.
For clarity, Algorithm~\ref{alg:knapsack} summarizes the main steps of the proposed Knapsack-based stratified sampling strategy.

\begin{algorithm}[htb]
\caption{Knapsack-based Stratified Sampling}
\label{alg:knapsack}
\fontsize{8pt}{8.8pt}\selectfont
\begin{algorithmic}[1]
\State \textbf{Input:} opinions $R$, sample size $N$, dimensions $D$, probabilities $P(c\mid r,d)$, relevance $\mathrm{rel}(r)$
\State \textbf{Output:} subset $S$
\State $S \gets \emptyset$
\ForAll{$d \in D,\; c \in C(d)$}
    \State compute $P(c\mid d)$, $N_{c\mid d}\gets \mathrm{round}(N\cdot P(c\mid d))$, $n_{c\mid d}\gets 0$
\EndFor
\State sort $R$ by decreasing $\mathrm{rel}(r)$
\ForAll{$r \in R$}
    \If{$|S|=N$} \State \textbf{break} \EndIf
    \State feasible $\gets$ true
    \ForAll{$d \in D$}
        \State $c^*\gets \arg\max_{c\in C(d)} P(c\mid r,d)$
        \If{$n_{c^*\mid d}+1>N_{c^*\mid d}$}
            \State feasible $\gets$ false; \textbf{break}
        \EndIf
    \EndFor
    \If{feasible}
        \State $S\gets S\cup\{r\}$
        \ForAll{$d \in D$}
            \State $c^*\gets \arg\max_{c\in C(d)} P(c\mid r,d)$
            \State $n_{c^*\mid d}\gets n_{c^*\mid d}+1$
        \EndFor
    \EndIf
\EndFor
\State \Return $S$
\end{algorithmic}
\end{algorithm}

The \texttt{Knapsack Kullback-Leibler} (\texttt{Knapsack-KL} or \texttt{KL}) variant enhances the representativeness of the selected subset by explicitly penalizing distributional drift. At each iteration, a small feasibility-restricted set of candidates is constructed, consisting only of elements that satisfy the same constraints enforced by the Knapsack formulation. Among these admissible candidates, the next element is selected by minimizing a scoring function that balances distributional fidelity with semantic relevance. Distributional fidelity is quantified through the Kullback-Leibler (KL) divergence between the empirical distributions of the candidate-augmented subset and those of the original corpus, while relevance is measured through the relevance score. Let \(x\) denote a candidate evaluated at iteration \(i\). For each such \(x\), we consider the hypothetical subset obtained by adding \(x\) to the current selection and compute the induced empirical distributions used in the KL term. The trade-off between relevance and distributional alignment is governed by a scheduling coefficient \(\alpha(i)\), which starts near \(1\) and decreases linearly toward \(0\), thereby prioritizing highly relevant items in early iterations and increasingly favoring candidates that improve distributional consistency in later stages. The scoring function for candidate \(x\) at iteration \(i\) is
\begin{equation}
f_{\text{score}}(x, i)
= \alpha(i)\, D_{\mathrm{KL}}(x)
+ \bigl(1 - \alpha(i)\bigr)\, \bigl(1 - \text{relevance\_score}(x)\bigr).
\label{eq:score}
\end{equation}

To make the KL component in~\eqref{eq:score} explicit, let \(P_d\) denote the original corpus distribution over dimension \(d\), and let \(Q_d^{+x}\) denote the empirical distribution obtained by augmenting the current subset with \(x\). The averaged KL divergence is
\begin{equation}
D_{\mathrm{KL}}(x)
= \frac{1}{D}
\sum_{d=1}^{D}
\mathrm{KL}\!\left(Q_d^{+x} \,\Vert\, P_d\right).
\label{eq:kl}
\end{equation}

This formulation ensures that each candidate is evaluated not only for its intrinsic relevance but also for its impact on the distributional structure of the evolving subset, while the constraint-filtered candidate set preserves the feasibility properties of the underlying Knapsack method. For clarity, Algorithm~\ref{alg:knapsackkl} summarizes the main steps of the proposed Knapsack-KL strategy.

\begin{algorithm}[htb]
\caption{Knapsack-KL Stratified Sampling}
\label{alg:knapsackkl}
\fontsize{8pt}{8.8pt}\selectfont
\begin{algorithmic}[1]
\State \textbf{Input:} opinions $R$, sample size $N$, dimensions $D$, probabilities $P(c\mid r,d)$, relevance $\mathrm{rel}(r)$
\State \textbf{Output:} subset $S$
\State $S \gets \emptyset$
\ForAll{$d \in D,\; c \in C(d)$}
    \State compute $P(c\mid d)$, $N_{c\mid d}\gets \mathrm{round}(N\cdot P(c\mid d))$, $n_{c\mid d}\gets 0$
\EndFor
\For{$i=1$ to $N$}
    \State build admissible set $A \subseteq R\setminus S$
    \ForAll{$x \in A$}
        \State compute $Q_d^{+x}$ and $D_{\mathrm{KL}}(x)$ using Eq.~\eqref{eq:kl}
        \State compute $f_{\mathrm{score}}(x,i)$ using Eq.~\eqref{eq:score}
    \EndFor
    \State $x^* \gets \arg\min_{x\in A} f_{\mathrm{score}}(x,i)$
    \State $S \gets S\cup\{x^*\}$
    \ForAll{$d \in D$}
        \State $c^*\gets \arg\max_{c\in C(d)} P(c\mid x^*,d)$
        \State $n_{c^*\mid d}\gets n_{c^*\mid d}+1$
    \EndFor
\EndFor
\State \Return $S$
\end{algorithmic}
\end{algorithm}

Finally, the \texttt{{Kernel Density Estimation} (\texttt{KDE})} strategy directly approximates the empirical distributions of class probabilities for both the full corpus and the evolving subset. In practice, these distributions are estimated using histogram-based density approximations over the probability space for each class and dimension. At each iteration, the method evaluates the scoring function for \emph{every element in the dataset}—rather than restricting attention to a feasibility-filtered candidate set as in the Knapsack-KL method. The next item selected is therefore the one that minimizes a scoring function, which jointly accounts for distributional alignment and semantic relevance. As in the Knapsack-KL approach, it uses the same scoring function~\eqref{eq:score}; the relevance component ensures that highly informative elements are preferred when multiple options yield similar distributional improvements. Although this greedy procedure is computationally heavier, because it repeatedly recomputes histogram estimates and evaluates the scoring function across the entire dataset, it yields the most faithful reconstruction of the original multi-dimensional probability distributions and thus produces the most semantically representative subsets. For conciseness, pseudocode for \texttt{KDE} is omitted because the algorithm follows the same iterative scoring-and-selection pattern as \texttt{Knapsack-KL}, with the main difference lying in its global candidate evaluation strategy.

Taken together, these strategies allow us to trade off computational cost and representational fidelity. Knapsack is fast and suitable for large-scale deployments, while Knapsack-KL and KDE provide increasingly precise control over distributional alignment and semantic preservation, at higher computational cost. In all cases, the selected subset is designed to maintain the diversity and structure of the opinion space under a tight token budget, enabling the LLM to generate summaries that remain faithful to the underlying corpus. 

From a computational perspective, the three strategies exhibit different practical costs. \texttt{Knapsack} is the lightest, since it relies on relevance-based ordering and simple feasibility checks over the target class counts. \texttt{Knapsack-KL} is more expensive because, at each iteration, it evaluates admissible candidates by recomputing their impact on the distributional divergence from the original corpus. \texttt{KDE} is the most computationally demanding strategy, as it evaluates all remaining candidates and repeatedly updates distributional estimates over the full candidate set. In practice, this additional overhead is offset by the substantial reduction in LLM input size, making the trade-off favorable in token-constrained summarization scenarios.

\section{Experimental Results}
\label{sec:experiments}

We evaluate the proposed framework on three heterogeneous datasets spanning product feedback, hotel reviews, and political discussion. All corpora are enriched with facet annotations (Section~\ref{sec:multi-classification}) and summarized using a unified prompt. Our stratified sampling approach is compared against a baseline that feeds the full opinion set to the model.

The first dataset consists of \textit{Amazon electronics reviews}\footnote{\url{https://github.com/SCAlabUnical/UserReviewDatasets}}, comprising roughly 10k entries across 100 products. Each review includes structured fields such as star rating, title, free-text content, verification status, reviewer location, and timestamp, enabling fine-grained analysis of usability, reliability, and product satisfaction.
Here, summaries aim to surface the main strengths and weaknesses of each product as perceived by users.

For hotel-related feedback, we rely on a \textit{Tripadvisor collection} built from \textit{hotels in New York}\footnote{\url{https://github.com/SCAlabUnical/UserReviewDatasets}}, including star ratings, titles, dates of stay, and metadata such as hotel coordinates, sub-ratings for service and location. This dataset provides rich, aspect-oriented evaluations typical of hospitality reviews.
Summaries synthesize guests’ reported positives and negatives across service, comfort, cleanliness, and overall experience.

To extend the evaluation beyond product and service reviews, we incorporate political opinions from the \textit{USA2024 X/Twitter dataset}\footnote{\url{https://github.com/sinking8/usc-x-24-us-election}}
, which tracks discourse surrounding the 2024 U.S.\ presidential election. The collection contains millions of posts gathered before election day (November~5,~2024) via a headless-browser scraper and filtered through election-related keywords and candidate-specific hashtags. For our experiments, we extract a subset of posts authored by users whose self-reported location corresponds to one of seven key swing states (e.g., Arizona). Summaries here condense the main arguments and justifications expressed by users supporting each candidate.

All summaries are generated using \textit{GPT-5} via API, with dataset-specific prompts that incorporate the facet information relevant to each domain (e.g., product strengths and weaknesses, hotel-service aspects, and reasons for political support). For each dataset, the same prompt template is used in two conditions: $(i)$ supplying all opinions in raw form, and $(ii)$ supplying only the stratified sample, enriched with multidimensional annotations (sentiment, topics, emotions, and any domain-specific facets). Although modern LLMs support increasingly wider context windows, using very large inputs can still degrade summary quality as the context grows~\cite{yuan2024lv}, which further motivates the need for compact, semantically representative sampling. The implementation of the proposed framework, including the sampling strategies, is publicly available at \url{https://github.com/SCAlabUnical/semantic-preserving-summarization}.

\subsection{Step-by-Step Operation on TripAdvisor Hotel Reviews}

For illustration, we demonstrate our framework on a New York hotel from the Tripadvisor dataset (anonymized for privacy). Below, we report one representative positive review and one representative negative review. For each opinion, our framework detects the sentiment, assigns a topic label, and identifies the expressed emotion. These examples show how multidimensional classification enriches raw feedback, enabling the construction of a compact and balanced sample for summarization.

\begin{tcolorbox}[breakable=true,boxsep=2pt,left=2mm,right=1mm,top=1mm,bottom=1mm,
sharp corners, colback=lightyellow, colframe=black, boxrule=0.5pt,
fontupper=\fontsize{8pt}{10pt}\selectfont]

\texttt{Hotel (h)} = "\textit{[Anonymized] Hotel}", 
\texttt{Location} = "New York, NY", 
\texttt{Description (d)} = "Modern hotel near Times Square with on-site dining and business services."

\vspace{0.15cm}

\texttt{Example of positive review (pr)} = 
\{\textit{title (tr)}: "Great Stay Thanks to Sandra", 
\textit{text (t)}: "The hotel is well-located and the highlight was the kindness of receptionist Sandra during check-in. The room had a lovely city view, comfortable bed, and was very clean...", 
\textit{sentiment (s)}: \texttt{Positive},
\textit{topics (t)}: \texttt{Sandra's Excellent Service, Excellent Hotel Stay}, 
\textit{emotions (e)}: \texttt{Joy}\}

\vspace{0.15cm}

\texttt{Example of negative review (nr)} =
\{\textit{title (tr)}: "Terrible Experience", 
\textit{text (t)}: "Poor service, dirty room, lack of amenities, and uninterested staff. Charged a service fee for amenities not provided. I advise others to avoid this hotel near Times Square; I’m seeking a refund.", 
\textit{sentiment (s)}: \texttt{Negative},
\textit{topics (t)}: \texttt{Poor Hotel Service, Refund Issues}, 
\textit{emotions (e)}: \texttt{Anger/Sadness}\}
\end{tcolorbox}

\begin{figure}[htb!]
  \centering
  \subfigure[Class distributions (sentiment, topics, emotion) for the full dataset (top) and for the sample (bottom).\label{fig:distr}]
  {\includegraphics[width=0.47\linewidth]{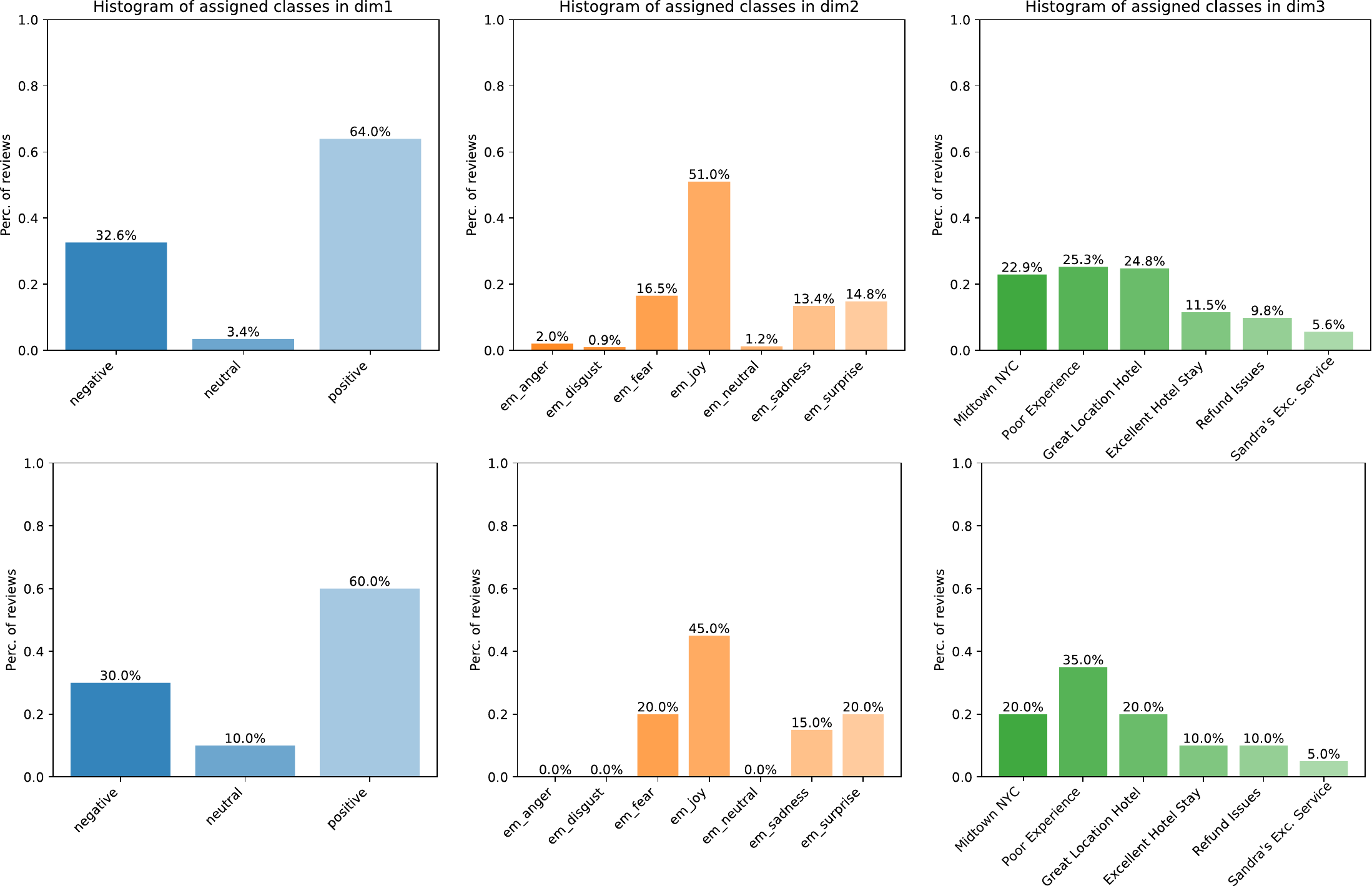}}
  \hspace{0.1cm}  
   \subfigure[Counts and density of sentiment (positive/negative) for full dataset (top) and sample (bottom).\label{fig:distr2}]  {\includegraphics[width=0.47\linewidth]{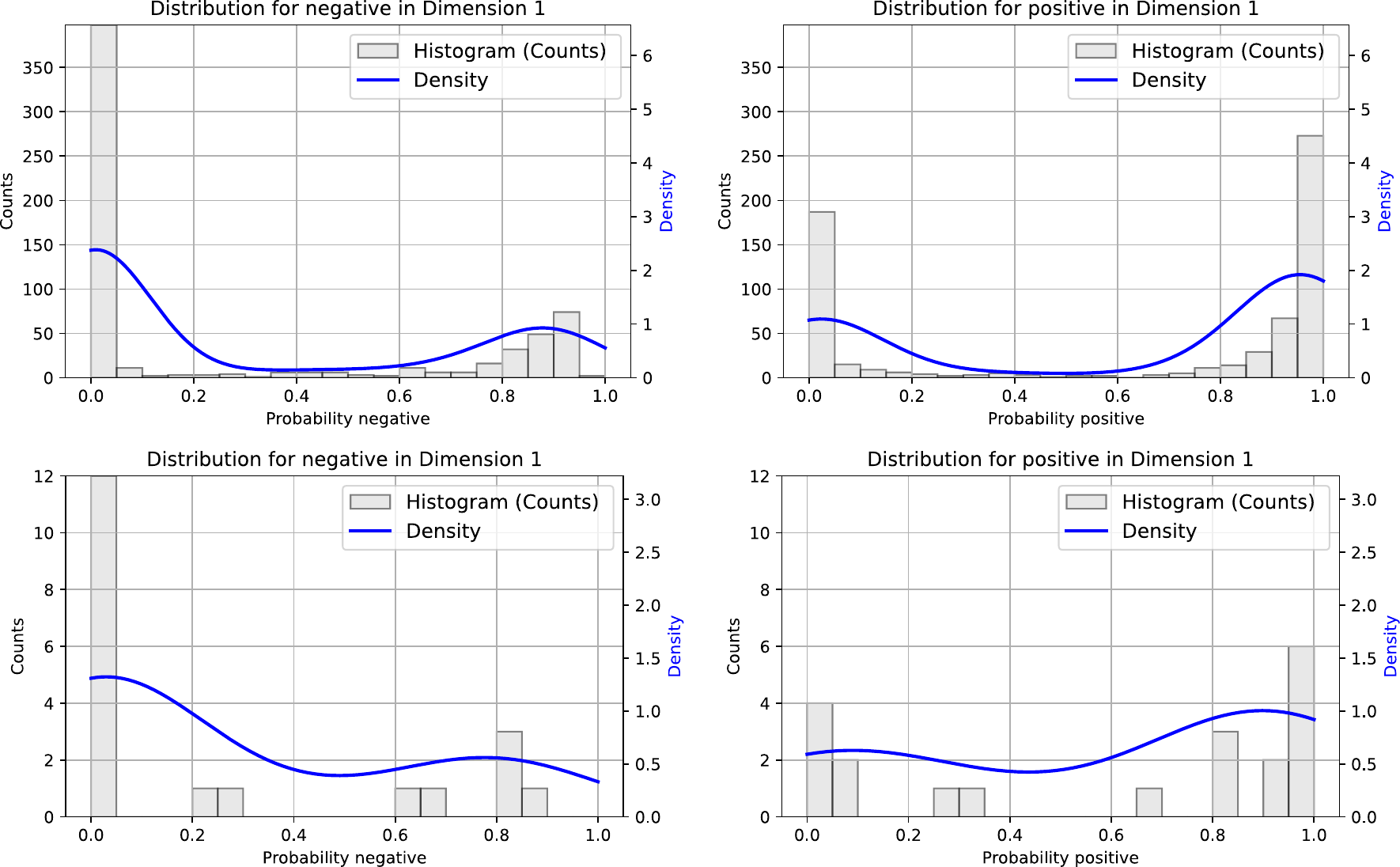}}
  \hspace{0.1cm}
  \subfigure[Topic clusters from BERTopic visualized with UMAP; ‘X’ marks indicate selected sample reviews.\label{fig:clust}]
{\includegraphics[width=0.7\linewidth]{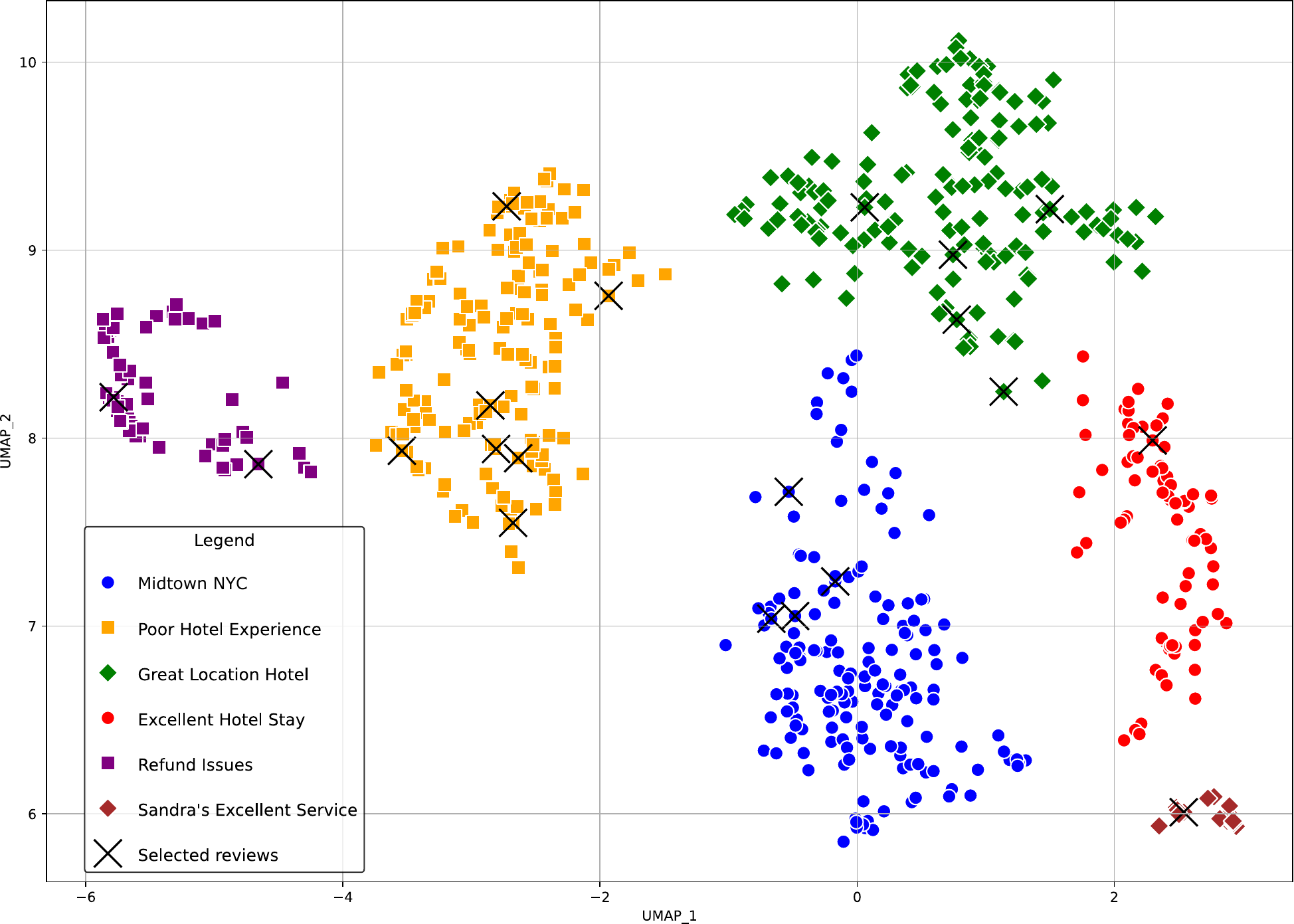}}  
  \label{fig:review_topics}
\caption{Class distributions for the full set of reviews of the example hotel and for the balanced sample (N = 20).}
\end{figure}

Among the advanced stratified samplers considered, in this example we adopt the \texttt{Kullback–Leibler}–based (\texttt{KL}) strategy to obtain a compact subset of \texttt{N}=20 reviews while preserving the semantic structure of the full corpus. As shown in Figure~\ref{fig:distr}, the class distributions for sentiment, topics, and emotion in the selected sample closely mirror those of the complete dataset. In addition to matching aggregate class frequencies, the sampler maintains the shape of the probability distribution within each dimension. Figure~\ref{fig:distr2} shows this for sentiment, where the distribution over the positive and negative classes remains consistent between the full corpus and the 20-review subset. The same distributional coherence is observed across all remaining dimensions. Finally, Figure~\ref{fig:clust} illustrates that the selected reviews---marked with an “X”---are spread across the topic clusters extracted through BERTopic and visualized via UMAP, demonstrating broad thematic coverage within the compact subset. Although this example uses the KL sampler, Knapsack and KDE exhibit similarly balanced, distribution-preserving behavior at the corpus level, but select different individual reviews, which naturally leads to high-quality yet distinct summaries.

To verify that a compact subset can still yield a high-quality summary, we compared summaries generated from the KL stratified sample with those obtained by providing the full set of hotel reviews to the LLM. Using ModernBERT embeddings, the summary produced from only 20 reviews reached a cosine similarity of about 0.85 to the full-context summary, whereas random sampling needed far larger subsets to achieve a similar level of semantic alignment. Token usage showed an even sharper contrast: summarizing the full set of hotel reviews required over 54,000 input tokens, whereas the stratified sample of 20 reviews required roughly 1500 tokens—more than a 97\% reduction—while maintaining high semantic fidelity. The following section provides a detailed quantitative analysis of these effects across all three datasets.

\subsection{Aggregate Metrics and Comparative Analysis}
\label{sec:aggregate_evaluation}

We evaluate the proposed framework along three main dimensions: (i) how well the sampled subsets preserve the topical structure of the full corpus (\emph{topic coverage}); (ii) how close the summaries generated from these subsets are to summaries obtained from all available opinions (\emph{content similarity}); and (iii) how many input tokens the LLM must process to generate the summary (\emph{token usage}). Results are averaged across all instances in the three datasets (Amazon products, Tripadvisor hotels, X/Twitter political posts), and all stratified sampling strategies are evaluated against a random baseline.

Compared with our previous conference work~\cite{ecml-pkdd-marozzo-2025}, the present study adopts a partially different evaluation protocol. In the conference version, the analysis focused primarily on summary quality through lexical, semantic, and qualitative measures, including TF-IDF, ROUGE, sentence-embedding similarities, BLANC, LLM-based evaluation, and human assessment. In contrast, the goal of the current paper is to compare multiple sampling strategies under strict token constraints and across heterogeneous domains. For this reason, we employ evaluation measures that are more directly aligned with this objective, namely topic coverage, summary-level cosine similarity, and token usage, which together quantify how well the selected subsets preserve the semantic structure of the full corpus while maintaining summarization efficiency.

\begin{figure}[b!]
  \centering
  \subfigure[Amazon products.\label{fig:amazon_topic_coverage}]{
    \includegraphics[width=0.3\linewidth]{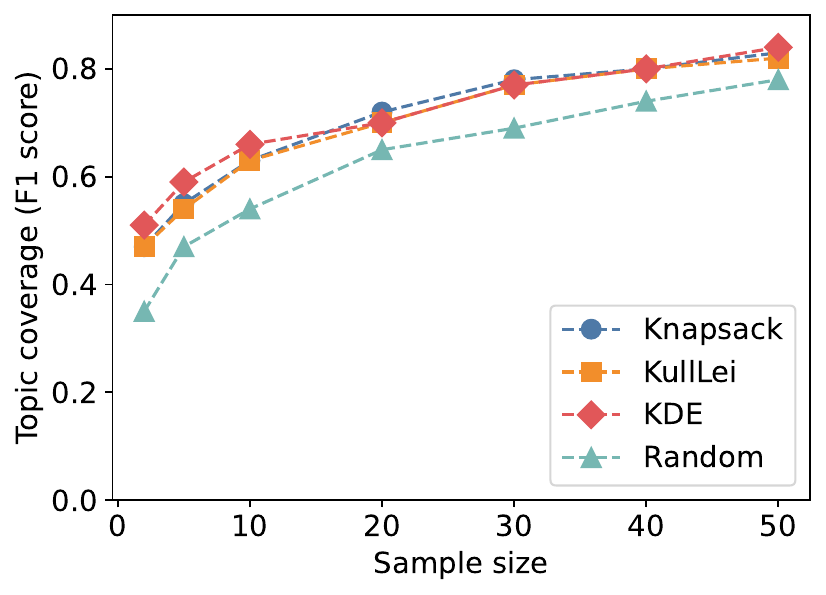}
  }
  \subfigure[Tripadvisor hotels.\label{fig:tripadvisor_topic_coverage}]{
    \includegraphics[width=0.3\linewidth]{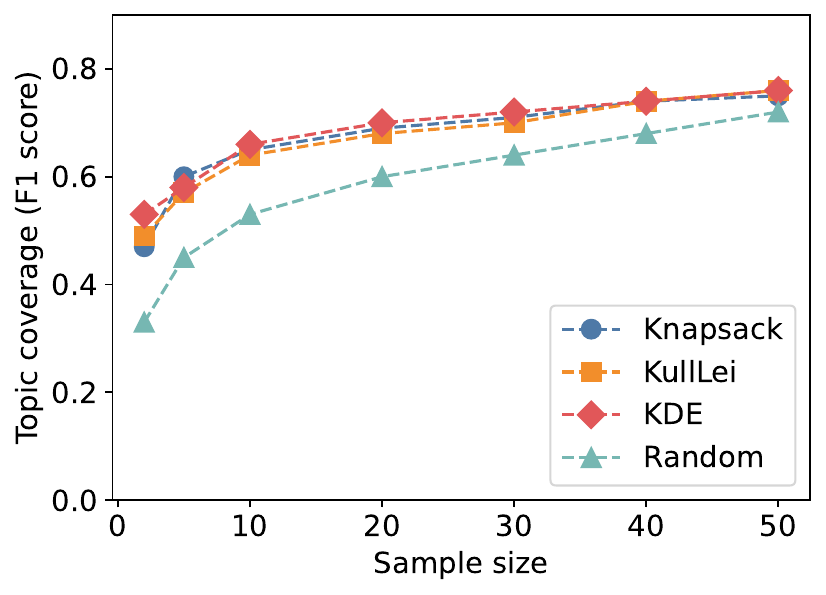}
  }  \subfigure[X/Twitter posts.\label{fig:twitter_topic_coverage}]{
    \includegraphics[width=0.3\linewidth]{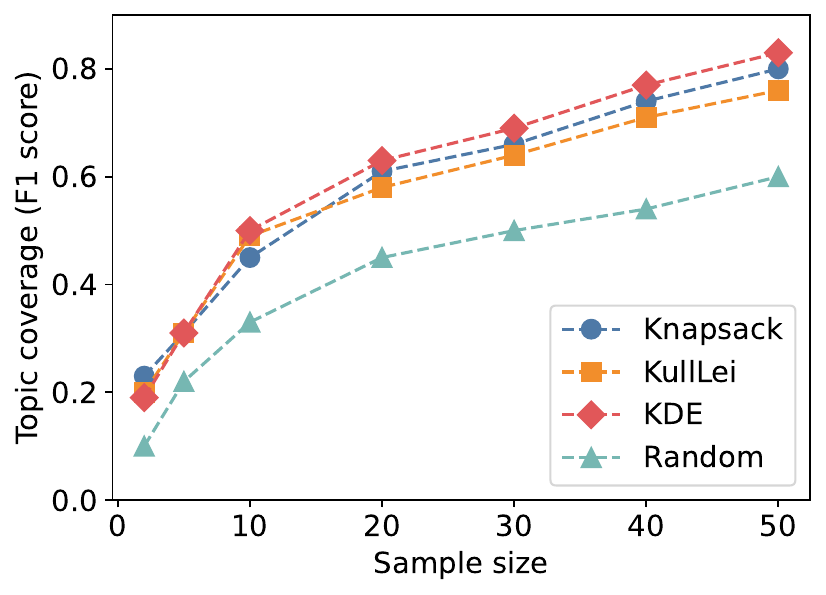}
  }  
  \caption{Topic coverage (F1 score) vs. sample size for different sampling strategies.}
  \label{fig:topic_coverage_all}
\end{figure}

\subsubsection{Topic coverage}
\label{sec:topic_coverage}

We first assess how well each sampling strategy preserves the topical structure of the full opinion set. For every instance in the Amazon, Tripadvisor, and X/Twitter datasets, we extract ground-truth topics from the entire corpus using \textit{ChatGPT-Thinking} with a structured prompt that groups themes into positive, neutral, and negative aspects. The same prompt is then applied to each sampled subset. Because topics are free-form text, we encode them with ModernBERT embeddings and compute semantic matches via cosine similarity. Standard set-based metrics (Precision, Recall, F1, Jaccard) are derived from these matches. The results (Figures \ref{fig:amazon_topic_coverage}–\ref{fig:twitter_topic_coverage}) show a consistent and clear trend: every stratified sampler substantially outperforms random selection, especially for small sample sizes. On Amazon, at $N$=20, stratified methods achieve F1 scores of 0.70–0.72, compared with 0.65 for random. Tripadvisor exhibits a similar gap, with stratified samplers reaching 0.68–0.70 versus 0.60 for random. The contrast is even stronger on X/Twitter---where topics are sparse and politically polarized---yielding 0.58–0.63 for the advanced samplers but only 0.45 for random. These results show that stratified selection retains far more topical diversity than random sampling across domains because it deliberately selects opinions so as to cover the main topics from the very beginning, rather than relying on chance to approximate the corpus structure. As sample size increases beyond $N$=20, all methods improve, but the relative ordering remains stable: stratified samplers always dominate random selection, and the performance gap narrows only gradually.

\subsubsection{Content similarity}
\label{sec:content_similarity}

We next assess how closely the summaries generated from sampled subsets match those produced from the full opinion set. For each instance, we first generate a reference summary using all available opinions, then create sample-based summaries of increasing size using each sampling strategy. Both summaries are embedded with ModernBERT, and cosine similarity is computed between their embeddings. Empirically, values above 0.8 correspond to summaries that are semantically equivalent, differing mainly in style or phrasing.

\begin{figure}[htb!]
  \centering
  \subfigure[Amazon products.\label{fig:amazon_cosine}]{
    \includegraphics[width=0.3\linewidth]{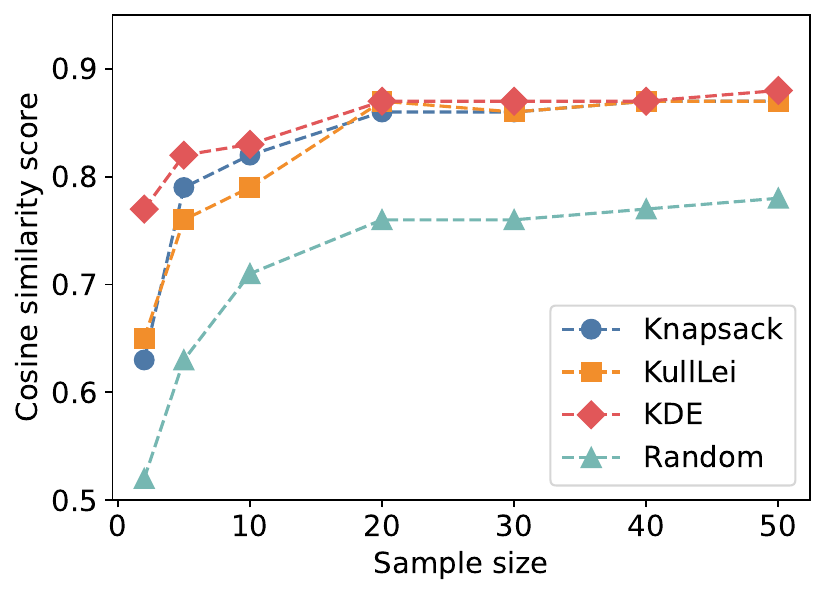}
  }
  \subfigure[Tripadvisor hotels.\label{fig:tripadvisor_cosine}]{
    \includegraphics[width=0.3\linewidth]{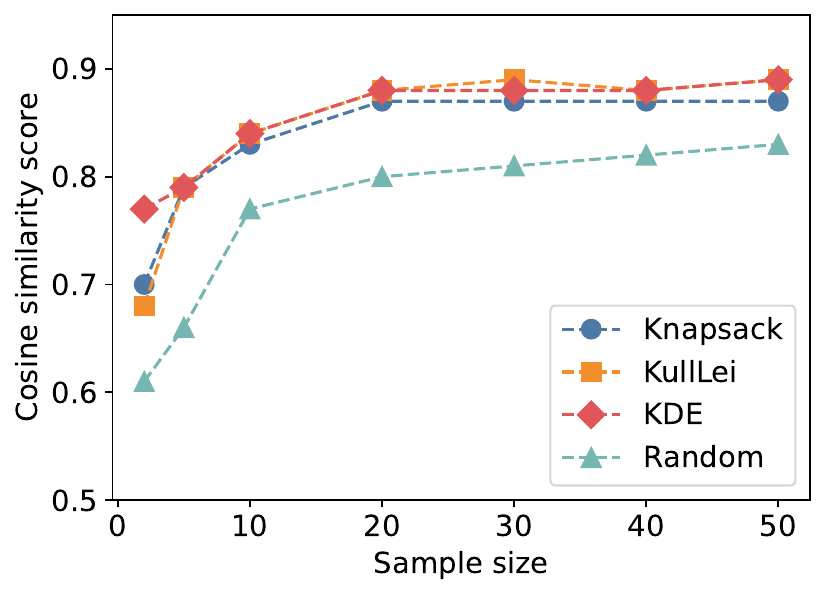}
  }  \subfigure[X/Twitter posts.\label{fig:twitter_cosine}]{
    \includegraphics[width=0.3\linewidth]{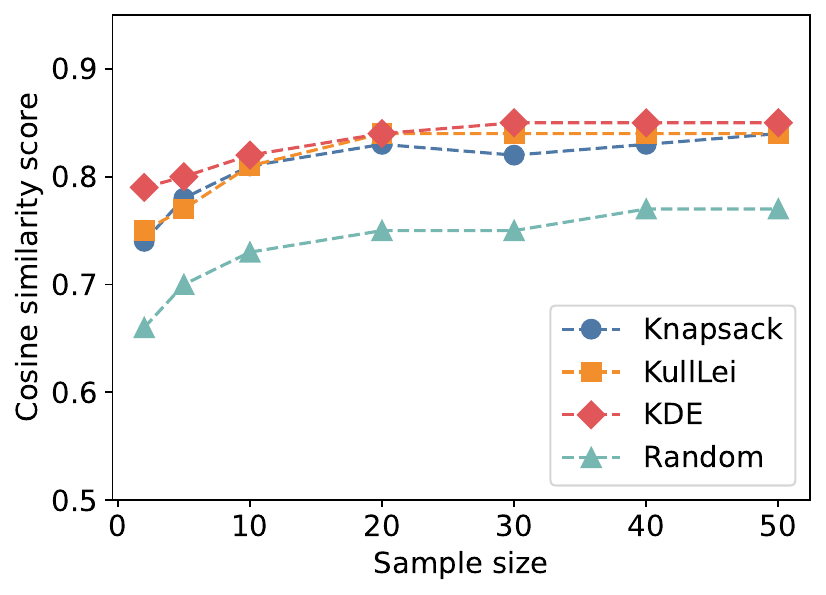}
  }  
  \caption{Summary-level cosine similarity vs.\ sample size for different sampling strategies.}
  \label{fig:summary_similarity_all}
\end{figure}

Across all datasets, stratified samplers consistently outperform random selection (Figures~\ref{fig:amazon_cosine}-\ref{fig:twitter_cosine}). On Amazon, at $N$=20, stratified sampling reaches cosine similarities of 0.86–0.87 (vs.\ 0.76 for random); on Tripadvisor, the same sample size yields 0.87–0.88 (vs.\ 0.80 for random). The advantage persists on X/Twitter, where stratified methods reach 0.83–0.84 at $N$=20, compared with 0.75 for random. KDE is particularly effective for very small samples: with only 5 opinions, KDE achieves 0.82 on Amazon, 0.79 on Tripadvisor, and 0.80 on X/Twitter---substantially higher than random’s 0.63, 0.66, and 0.70, respectively. As sample size increases beyond 20 opinions, cosine similarity plateaus around 0.85–0.88 across all stratified methods, indicating that additional opinions yield diminishing improvements once the main semantic content has been captured.

\subsubsection{Token usage}
\label{sec:token_usage_eval}

\begin{figure}[tb!]
  \centering
  \subfigure[Amazon products.\label{fig:amazon_token}]{
    \includegraphics[width=0.3\linewidth]{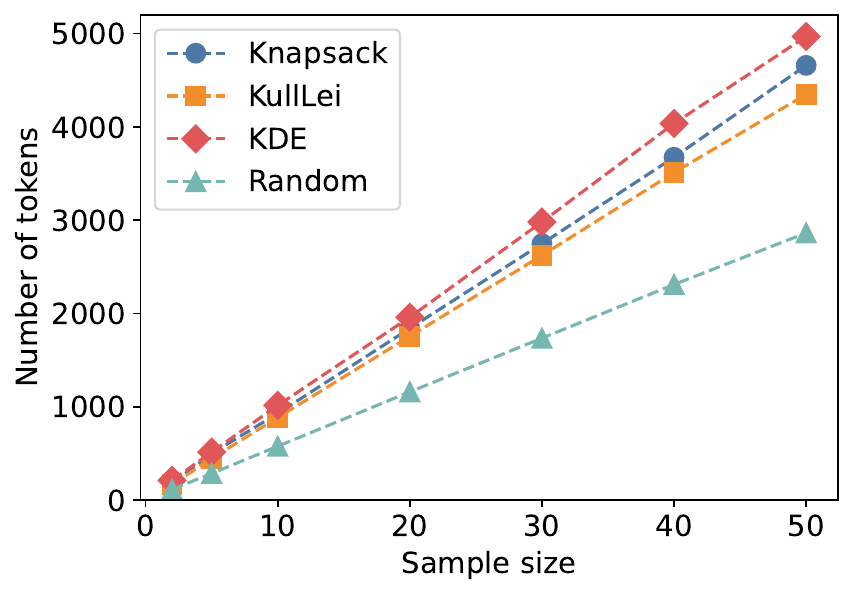}
  }
  \subfigure[Tripadvisor hotels.\label{fig:tripadvisor_token}]{
    \includegraphics[width=0.3\linewidth]{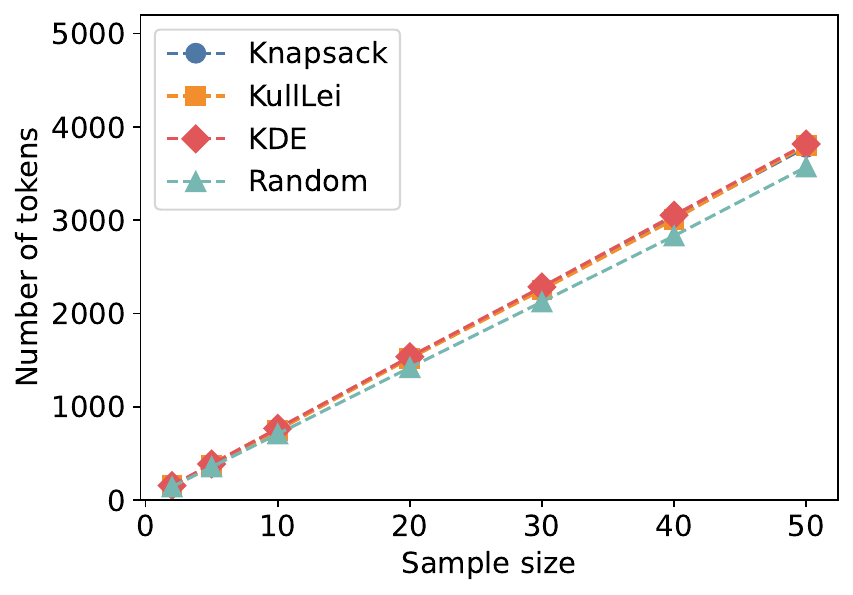}
  }  \subfigure[X/Twitter posts.\label{fig:twitter_token}]{
    \includegraphics[width=0.3\linewidth]{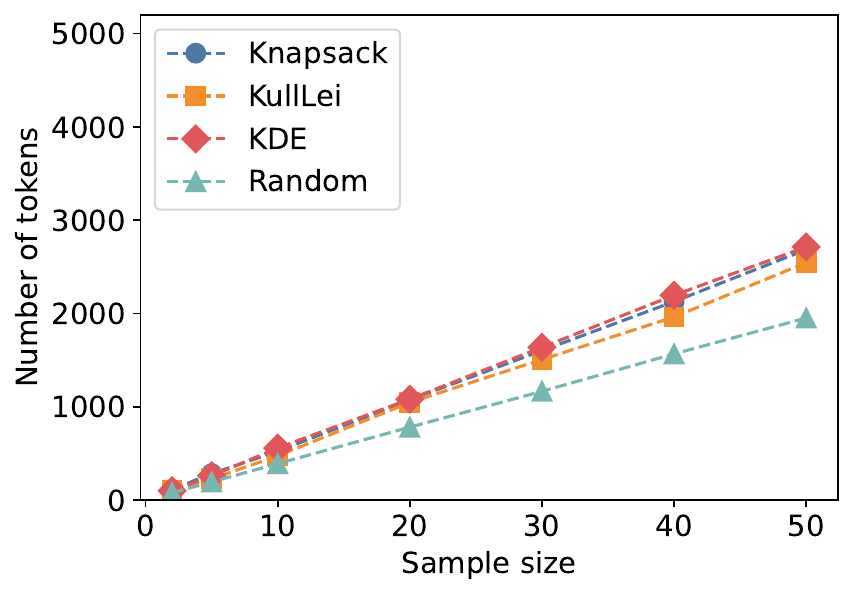}
  }  
  \caption{Token usage vs.\ sample size across sampling strategies and random selection.\label{fig:alg_token}}
\end{figure}

A central objective of our framework is to minimize the number of input tokens the LLM must process while still preserving the semantic content of the full corpus. Figures~\ref{fig:amazon_token}–\ref{fig:twitter_token} report the average number of tokens associated with samples of increasing size. As expected, token usage grows roughly linearly with the number of selected opinions. Stratified samplers generally use slightly more tokens than random selection at the same $N$, because they tend to select longer, information-dense opinions. For example, at $N$=20, stratified methods require 1,756–1,961 tokens on Amazon, 1,519–1,537 on Tripadvisor, and 1,047–1,082 on X/Twitter, compared with 1,160, 1,420, and 782 for random sampling.

However, when compared to the cost of processing the entire dataset, these values represent a dramatic reduction. For Amazon products, full-corpus summaries typically exceed 28,000 tokens, whereas stratified samples achieve cosine similarity above 0.83 with only 500–2,000 tokens (5–20 reviews), i.e., roughly 1–2\% of the tokens required by the full set. The savings are similar on Tripadvisor---where hotels generally have even more reviews---and on X/Twitter, where a dataset corpus can exceed 100,000 tokens. Crucially, for any fixed token budget, stratified samplers consistently achieve higher semantic fidelity than random selection: even with very small budgets, stratified subsets reach high cosine-similarity scores that random sampling can match only with substantially larger inputs. Comparable trends appear across all domains, where balanced sampling delivers near–full-context semantic quality at a fraction of the computational cost.

\subsubsection{Comparison of Sampling Strategies and Ablation on Semantic Dimensions}
\label{sec:final_comparison}

Table~\ref{tab:final_comparison} reports the average performance of the stratified sampling strategies at $N$=20 across all datasets. All methods clearly outperform random selection, confirming that distribution-aware sampling substantially improves both topic preservation and summary fidelity under tight token budgets. Knapsack offers a strong balance between coverage, summary quality, and speed, making it a practical choice when computational efficiency is a priority.
Knapsack–KL performs similarly to Knapsack, with only marginal gains or losses depending on the dataset, indicating that KL regularization provides limited additional benefit.
KDE achieves the highest topic coverage and cosine similarity, especially in semantically heterogeneous datasets, but incurs the highest computational cost due to repeated distributional updates. 
Overall, stratified sampling yields large token savings---typically 97.5\% fewer tokens than summarizing the full corpus---while producing summaries that remain semantically faithful to those obtained under full-context processing. KDE is the most effective when semantic preservation is paramount, whereas Knapsack provides the best quality–efficiency trade-off.

\begin{table}[htb!]
\centering
\setlength{\tabcolsep}{7pt} 
\fontsize{8pt}{9pt}\selectfont
\begin{tabular}{lccccc}
\toprule
\textbf{Algorithm} &
\textbf{Topic (F1)} &
\textbf{Cosine sim.} &
\textbf{Avg.\ tokens} &
\textbf{Token reduction} &
\textbf{Speed} \\
\midrule
Knapsack      & 0.67 & 0.85 & 1{,}479 & $\sim$97.5\% & Fast \\
Knapsack-KL  & 0.65 & 0.86 & 1{,}440 & $\sim$97.6\% & Medium \\
KDE           & 0.68 & 0.87 & 1{,}526 & $\sim$97.4\% & Slowest \\
Random        & 0.57 & 0.77 & 1{,}120 & $\sim$98.1\% & Fastest \\
\bottomrule
\end{tabular}
\caption{Average performance of sampling strategies at $N$=20 across Amazon, Tripadvisor, and X/Twitter.}
\label{tab:final_comparison}
\end{table}

To better understand the contribution of the multidimensional representation to the quality of the selected subsets, we performed a focused ablation study by progressively varying the semantic dimensions used during stratified sampling. In particular, we compared three configurations: \textit{Sentiment only}, \textit{Sentiment + Topic}, and \textit{Sentiment + Topic + Emotion}. The ablation adopts the same evaluation setting used in Table~\ref{tab:final_comparison}, namely $N=20$ across all datasets, and uses KDE as the reference sampling strategy, since it achieved the best overall performance in the main comparison.

For each configuration, we measured topic coverage and summary-level cosine similarity with respect to the summary generated from the full corpus. Table~\ref{tab:ablation_dimensions} summarizes the results. As expected, the \textit{Sentiment only} configuration yields the weakest performance, since polarity alone is insufficient to capture the semantic diversity of opinionated corpora. Adding \textit{Topic} information produces a clear improvement, confirming that thematic coverage is essential for selecting representative opinions. The full \textit{Sentiment + Topic + Emotion} configuration achieves the best overall results, showing that emotion provides an additional complementary signal for preserving nuance and viewpoint diversity in the generated summaries.

Overall, these findings indicate that the effectiveness of the proposed framework depends not only on the sampling mechanism itself, but also on the richness of the semantic representation used to guide selection. In particular, they support the claim that multidimensional classification is a key ingredient for obtaining compact yet semantically faithful summaries under strict token constraints.

\begin{table}[htb!]
\centering
\setlength{\tabcolsep}{8pt}
\fontsize{8pt}{9pt}\selectfont
\begin{tabular}{lcc}
\toprule
\textbf{Configuration} & \textbf{Topic F1} & \textbf{Cosine sim.} \\
\midrule
Sentiment only              & 0.56 & 0.80 \\
Sentiment + Topic           & 0.66 & 0.85 \\
Sentiment + Topic + Emotion & 0.68 & 0.87 \\
\bottomrule
\end{tabular}
\caption{Ablation study on the semantic dimensions used during stratified sampling, using KDE under the same setting as Table~\ref{tab:final_comparison} ($N=20$ across all datasets).}
\label{tab:ablation_dimensions}
\end{table}

\section{Conclusions}
\label{sec:conclusions}

This work addressed the problem of generating token-efficient summaries of opinionated text while preserving semantics and viewpoint diversity across domains such as e-commerce, hospitality, and political discourse. We proposed a corpus-level framework that first applies multidimensional classification (sentiment, topics, emotion, and optional domain-specific facets) and then uses stratified sampling strategies to select a compact, facet-balanced subset of opinions, which is finally summarized by an LLM through tailored prompts. Experiments on Amazon product reviews, Tripadvisor hotel reviews, and political discussions on X/Twitter show that the proposed methods consistently outperform random sampling in topic coverage and summary-level cosine similarity, while reducing token usage by one to two orders of magnitude compared with summarizing the full corpus. Among the evaluated strategies, KDE achieves the highest semantic fidelity, whereas Knapsack and Knapsack-KL provide more favorable trade-offs between summary quality and computational cost.

Future work will extend the framework in several directions. We plan to evaluate it on additional and multilingual opinion sources to assess its robustness across domains and languages, and to explore interactive variants in which analysts can prioritize specific facets, such as safety-related issues, minority viewpoints, or emerging concerns. We also aim to investigate tighter integration with retrieval-augmented generation, combining corpus-level balance with fine-grained evidence retrieval to further improve the faithfulness and contextual grounding of the generated summaries.

\section*{Acknowledgements}
This work was supported by the research project “INSIDER: INtelligent ServIce Deployment for advanced cloud-Edge integRation” granted by the Italian Ministry of University and Research (MUR) within the PRIN 2022 program and European Union - Next Generation EU (grant n. 2022WWSCRR,  CUP H53D23003670006). It was also supported by the ``National Centre for HPC, Big Data and Quantum Computing", CN00000013 - CUP H23C22000360005, and by the ``FAIR – Future Artificial Intelligence Research" project - CUP H23C22000860006.

\bibliographystyle{plainnat}
\makeatletter
\renewcommand{\@biblabel}[1]{\hfill[#1]}  
\renewenvironment{thebibliography}[1]
     {\section*{\refname}
      \small                
      \begin{list}{\@biblabel{\@arabic\c@enumiv}}%
           {\usecounter{enumiv}\setlength{\itemsep}{0pt}  
            \setlength{\parsep}{0pt}
            \setlength{\leftmargin}{\bibindent}}%
      \sloppy
     }
     {\end{list}}
\makeatother
\bibliography{references}

@inproceedings{carichon2023topically,
  title={Topically diversified summarization of customer reviews},
  author={Carichon, Florian and Caporossi, Gilles},
  booktitle={Proceedings of the 6th International Conference on Natural Language and Speech Processing (ICNLSP 2023)},
  pages={178--191},
  year={2023}
}

@inproceedings{jiang2023large,
  title={Large-scale and multi-perspective opinion summarization with diverse review subsets},
  author={Jiang, Han and Wang, Rui and Wei, Zhihua and Li, Yu and Wang, Xinpeng},
  booktitle={Findings of the Association for Computational Linguistics: EMNLP 2023},
  pages={5641--5656},
  year={2023}
}

@inproceedings{zhang2015clustering,
  title={Clustering sentences with density peaks for multi-document summarization},
  author={Zhang, Yang and Xia, Yunqing and Liu, Yi and Wang, Wenmin},
  booktitle={Proceedings of the 2015 conference of the north american chapter of the association for computational linguistics: Human language technologies},
  pages={1262--1267},
  year={2015}
}

@inproceedings{carbonell1998use,
  title={The use of MMR, diversity-based reranking for reordering documents and producing summaries},
  author={Carbonell, Jaime and Goldstein, Jade},
  booktitle={Proceedings of the 21st annual international ACM SIGIR conference on Research and development in information retrieval},
  pages={335--336},
  year={1998}
}

@article{sharma2022automatic,
  title={Automatic text summarization methods: A comprehensive review},
  author={Sharma, Grishma and Sharma, Deepak},
  journal={SN Computer Science},
  volume={4},
  number={1},
  pages={33},
  year={2022},
  publisher={Springer}
}

@inproceedings{novelneeds2024,
author = {Huang, Shaoqin and Wang, Yue and Mo, Daniel Y. and Liu, Hai},
title = {Mining Novel Customer Needs from Online Product Review},
year = {2024},
isbn = {9798400718151},
publisher = {Association for Computing Machinery},
address = {New York, NY, USA},
url = {https://doi.org/10.1145/3686081.3686108},
doi = {10.1145/3686081.3686108},
booktitle = {Proceedings of the International Conference on Decision Science \& Management},
pages = {168–172},
numpages = {5},
location = {
},
series = {ICDSM '24}
}

@article{reviewsproductdev2020,
   title={Mining customer product reviews for product development: A summarization process},
   volume={132},
   ISSN={0957-4174},
   url={http://dx.doi.org/10.1016/j.eswa.2019.04.069},
   DOI={10.1016/j.eswa.2019.04.069},
   journal={Expert Systems with Applications},
   publisher={Elsevier BV},
   author={Hou, Tianjun and Yannou, Bernard and Leroy, Yann and Poirson, Emilie},
   year={2019},
   month=oct, pages={141–150} }

@InProceedings{product-comp-ds25,
      AUTHOR = {Cosentino, Cristian and Gunduz-Cure, Merve and Marozzo, Fabrizio and Ozturk-Birim, Sule},
      TITLE = {From Reviews to Results: Generative AI for Review-Driven Product and Service Comparisons},
      YEAR = {2025},
      BOOKTITLE = {28th International Conference on Discovery Science (DS2025)},
      PAGES = {78–93}
}

@article{Chengyao2025survey,
author = {Lv, Chengyao and Tang, Yiwen and Ao, Lian and Huang, Yanxia and Zhang, Simin and Fan, Junqing and Han, Wei},
year = {2025},
month = {10},
pages = {801-824},
title = {A survey of automatic text summarization: concepts, advances and future prospects},
volume = {28},
journal = {International Journal of Speech Technology},
doi = {10.1007/s10772-025-10215-y}
}

@inproceedings{10.1145/3477495.3532676,
author = {Kim Amplayo, Reinald and Brazinskas, Arthur and Suhara, Yoshi and Wang, Xiaolan and Liu, Bing},
title = {Beyond Opinion Mining: Summarizing Opinions of Customer Reviews},
year = {2022},
isbn = {9781450387323},
publisher = {Association for Computing Machinery},
address = {New York, NY, USA},
url = {https://doi.org/10.1145/3477495.3532676},
doi = {10.1145/3477495.3532676},
booktitle = {Proceedings of the 45th International ACM SIGIR Conference on Research and Development in Information Retrieval},
pages = {3447–3450},
numpages = {4},
location = {Madrid, Spain},
series = {SIGIR '22}
}

@InProceedings{ecml-pkdd-marozzo-2025,
      AUTHOR = {Belcastro, Loris and et al.},
      TITLE = {Balanced and Token-Efficient Summarization of User Reviews via Stratified Sampling and Large Language Models},
      YEAR = {2025},
      BOOKTITLE = {European Conference on Machine Learning and Principles and Practice of Knowledge Discovery in Databases (ECML PKDD)},
      PAGES = {290–306}
}

@inproceedings{zhang2019bertscore,
  title     = {BERTScore: Evaluating Text Generation with BERT},
  author    = {Zhang, Tianyi and Kishore, Varsha and Wu, Felix and Weinberger, Kilian Q. and Artzi, Yoav},
  booktitle = {ICLR},
  year      = {2020},
  eprint    = {1904.09675},
  archivePrefix = {arXiv},
  primaryClass  = {cs.CL},
  url       = {https://arxiv.org/abs/1904.09675}
}

@article{li2025coverage-fairness,
  title   = {Coverage-based Fairness in Multi-document Summarization},
  author  = {Li, Haoyuan and Zhang, Yusen and Zhang, Rui and Chaturvedi, Snigdha},
  journal = {arXiv preprint arXiv:2412.08795},
  year    = {2025},
  url     = {https://arxiv.org/abs/2412.08795}
}

@inproceedings{lewis2020rag,
  title     = {Retrieval-Augmented Generation for Knowledge-Intensive NLP Tasks},
  author    = {Lewis, Patrick and et al.},
  booktitle = {Advances in Neural Information Processing Systems (NeurIPS)},
  year      = {2020},
  url       = {https://arxiv.org/abs/2005.11401}
}

@article{gao2023rag_survey,
  title   = {Retrieval-Augmented Generation for Large Language Models: A Survey},
  author  = {Gao, Yunfan and et al.},
  journal = {arXiv preprint arXiv:2312.10997},
  year    = {2023},
  url     = {https://arxiv.org/abs/2312.10997}
}

@article{yuan2024lv,
  title={Lv-eval: A balanced long-context benchmark with 5 length levels up to 256k},
  author={Yuan, Tao and Ning, Xuefei and Zhou, Dong and Yang, Zhijie and Li, Shiyao and Zhuang, Minghui and Tan, Zheyue and Yao, Zhuyu and Lin, Dahua and Li, Boxun and others},
  journal={arXiv preprint arXiv:2402.05136},
  year={2024}
}

@article{zhang2024survey,
  title={Survey of transformers and towards ensemble learning using transformers for natural language processing},
  author={Zhang, Hongzhi and Shafiq, M Omair},
  journal={Journal of big Data},
  volume={11},
  number={1},
  pages={25},
  year={2024},
  publisher={Springer}
}

@article{roumeliotis2024llms,
  title={LLMs in e-commerce: a comparative analysis of GPT and LLaMA models in product review evaluation},
  author={Roumeliotis, Konstantinos I and Tselikas, Nikolaos D and Nasiopoulos, Dimitrios K},
  journal={Natural Language Processing Journal},
  volume={6},
  pages={100056},
  year={2024},
  publisher={Elsevier}
}

@article{sun2023text,
  title={Text classification via large language models},
  author={Sun, Xiaofei and Li, Xiaoya and Li, Jiwei and Wu, Fei and Guo, Shangwei and Zhang, Tianwei and Wang, Guoyin},
  journal={arXiv:2305.08377},
  year={2023}
}

@inproceedings{liu2005opinion,
  title={Opinion observer: analyzing and comparing opinions on the web},
  author={Liu, Bing and Hu, Minqing and Cheng, Junsheng},
  booktitle={14th Int. Conf. on World Wide Web},
  pages={342--351},
  year={2005}
}

@article{pang2002thumbs,
  title={Thumbs up? Sentiment classification using machine learning techniques},
  author={Pang, Bo and Lee, Lillian and Vaithyanathan, Shivakumar},
  journal={cs/0205070},
  year={2002}
}

@article{yang2019exploiting,
  title={Exploiting user experience from online customer reviews for product design},
  author={Yang, Bai and et al.},
  journal={Int. J. of Information Management},
  volume={46},
  pages={173--186},
  year={2019},
  publisher={Elsevier}
}

@article{zhang2014examining,
  title={Examining the influence of online reviews on consumers' decision-making: A heuristic--systematic model},
  author={Zhang, Kem ZK and Zhao, Sesia J and Cheung, Christy MK and Lee, Matthew KO},
  journal={Decision support systems},
  volume={67},
  pages={78--89},
  year={2014},
  publisher={Elsevier}
}

@inproceedings{lin2004rouge,
  title={Rouge: A package for automatic evaluation of summaries},
  author={Lin, Chin-Yew},
  booktitle={Text summarization branches out},
  pages={74--81},
  year={2004}
}

@article{grootendorst2022bertopic,
  title={BERTopic: Neural topic modeling with a class-based TF-IDF procedure},
  author={Grootendorst, Maarten},
  journal={arXiv:2203.05794},
  year={2022}
}

@inproceedings{cantini2024multi,
  title={Multi-dimensional Classification on Social Media Data for Detailed Reporting with Large Language Models},
  author={Cantini, Riccardo and Cosentino, Cristian and Marozzo, Fabrizio},
  booktitle={Int. Conf. on Artificial Intelligence Applications and Innovations},
  pages={100--114},
  year={2024}
}

@article{okonkwo2021chatbots,
  title={Chatbots applications in education: A systematic review},
  author={Okonkwo, Chinedu Wilfred and Ade-Ibijola, Abejide},
  journal={Computers and Education: Artificial Intelligence},
  volume={2},
  pages={100033},
  year={2021},
  publisher={Elsevier}
}

@article{egger2022topic,
  title={A topic modeling comparison between lda, nmf, top2vec, and bertopic to demystify twitter posts},
  author={Egger, Roman and Yu, Joanne},
  journal={Frontiers in sociology},
  volume={7},
  year={2022},
  publisher={Frontiers Media SA}
}

@article{info13010041,
AUTHOR = {Caldarini, Guendalina and Jaf, Sardar and McGarry, Kenneth},
TITLE = {A Literature Survey of Recent Advances in Chatbots},
JOURNAL = {Information},
VOLUME = {13},
YEAR = {2022},
NUMBER = {1},
ARTICLE-NUMBER = {41}
}

@article{ROUMELIOTIS2024100056,
title = {LLMs in e-commerce: A comparative analysis of GPT and LLaMA models in product review evaluation},
journal = {Natural Language Processing Journal},
volume = {6},
pages = {100056},
year = {2024},
issn = {2949-7191},
doi = {https://doi.org/10.1016/j.nlp.2024.100056},
url = {https://www.sciencedirect.com/science/article/pii/S2949719124000049},
author = {Konstantinos I. Roumeliotis and Nikolaos D. Tselikas and Dimitrios K. Nasiopoulos},
}

@article{anbumani2024enhancing,
  title={Enhancing sentiment analysis classification for amazon product reviews using CNN-sigTan-Beta activation function},
  author={Anbumani, P and Selvaraj, K},
  journal={Multimedia Tools and Applications},
  volume={83},
  number={19},
  pages={56719--56736},
  year={2024},
  publisher={Springer}
}

@article{10.1145/3615356,
author = {Perti, Ashwin and Sinha, Amit and Vidyarthi, Ankit},
title = {Cognitive Hybrid Deep Learning-based Multi-modal Sentiment Analysis for Online Product Reviews},
year = {2024},
issue_date = {August 2024},
address = {New York, NY, USA},
volume = {23},
number = {8},
issn = {2375-4699},
url = {https://doi.org/10.1145/3615356},
doi = {10.1145/3615356},
journal = {ACM Trans. Asian Low-Resour. Lang. Inf. Process.},
month = {aug},
articleno = {113},
numpages = {14}
}

@article{cantini2025harnessing,
  title={Harnessing prompt-based large language models for disaster monitoring and automated reporting from social media feedback},
  author={Cantini, Riccardo and et al.},
  journal={Online Social Networks and Media},
  volume={45},
  pages={100295},
  year={2025}
}

\end{document}